\documentclass{article}

\usepackage[final,nonatbib]{neurips_2018}

\usepackage[utf8]{inputenc} 
\usepackage[T1]{fontenc}    
\usepackage{booktabs}       
\usepackage{multirow}
\usepackage{amsfonts}       
\usepackage{nicefrac}       
\usepackage{microtype}      
\usepackage{graphicx}
\usepackage{inconsolata}    
\usepackage{caption}
\usepackage{subcaption}
\usepackage{listings}
\usepackage{placeins}
\usepackage[symbol]{footmisc}
\usepackage{tabularx}

\usepackage[numbers,sort&compress]{natbib}
\usepackage[hyphens]{url}           
\usepackage[hidelinks]{hyperref}    
\usepackage{glossaries}

\newcommand{\email}[1]{\small\texttt{\texorpdfstring{\href{mailto:#1}{#1}}{#1}}}

\renewcommand{\thefootnote}{\arabic{footnote}}

\newcommand{\astfootnote}[1]{%
\let\oldthefootnote=\thefootnote%
\setcounter{footnote}{0}%
\renewcommand{\thefootnote}{\fnsymbol{footnote}}%
\footnote{#1}%
\let\thefootnote=\oldthefootnote%
}

\lstdefinestyle{json}{
    breaklines=true,
    basicstyle=\small\ttfamily,
    numberstyle=\small\ttfamily,
    ,xleftmargin=2.5em,
    ,tabsize=2
}

\title{Document-Level Abstractive Summarization}

\author{%
  Gonçalo Raposo\\
  INESC-ID \\
  Instituto Superior Técnico \\
  Universidade de Lisboa \\
  \email{goncalo.cascalho.raposo@tecnico.ulisboa.pt} \\
  \And
  Afonso Raposo\\
  Instituto de Telecomunicações \\
  Instituto Superior Técnico \\
  Universidade de Lisboa \\
  \email{afonso.raposo@tecnico.ulisboa.pt} \\
  \And
  Ana Sofia Carmo\\
  Instituto de Telecomunicações \\
  Instituto Superior Técnico \\
  Universidade de Lisboa \\
  \email{ana.sofia.carmo@tecnico.ulisboa.pt} \\
}

\begin{document}
\newacronym{ae}{AE}{Autoencoder}

\newacronym{ats}{ATS}{Automatic Text Summarization}

\newacronym{ar}{AR}{Autoregressive}

\newacronym{bert}{BERT}{Bidirectional Encoder Representations from Transformers}

\newacronym{led}{LED}{Longformer-Encoder-Decoder}

\newacronym{llrd}{LLRD}{Layer-wise Learning Rate Decay}

\newacronym{nlp}{NLP}{Natural Language Processing}

\newacronym{s2s}{seq2seq}{sequence-to-sequence}

\newacronym{rnn}{RNN}{Recurrent Neural Network}

\newacronym{cca}{\textsc{Cca}}{chunked cross-attention}

\maketitle

\begin{abstract}
The task of automatic text summarization produces a concise and fluent text summary while preserving key information and overall meaning.
Recent approaches to document-level summarization have seen significant improvements in recent years by using models based on the Transformer architecture.
However, the quadratic memory and time complexities with respect to the sequence length make them very expensive to use, especially with long sequences, as required by document-level summarization.
Our work addresses the problem of document-level summarization by studying how efficient Transformer techniques can be used to improve the automatic summarization of very long texts.
In particular, we will use the arXiv dataset, consisting of several scientific papers and the corresponding abstracts, as baselines for this work. 
Then, we propose a novel retrieval-enhanced approach based on the architecture which reduces the cost of generating a summary of the entire document by processing smaller chunks. The results were below the baselines but suggest a more efficient memory a consumption and truthfulness.\astfootnote{This abstract was generated automatically using \textsc{LED}.}

\setcounter{footnote}{0}
\end{abstract}

\section{Introduction}

With the growth of publicly available text data, the summarization of such contents is essential for their usefulness. A text summary must convey important information from the original text and present a smaller, more manageable, size \cite{Radev2002}. The task of automatic text summarization produces a concise and fluent text summary while preserving key information and overall meaning \cite{Allahyari2017}.

Approaches to automatic text summarization can be divided into extractive and abstractive summarization. While the extractive approach produces a summary that is comprised entirely of excerpts from the original text, the abstractive approach generates an output that may contain content that is entirely original. Both approaches have seen significant improvements in recent years by using models based on the Transformer architecture \cite{Vaswani2017}. In particular, the fluency of these language models has allowed for state-of-the-art results for abstractive summarization \cite{Lewis2020,Raffel2020,Zhang2020}.

However, Transformers' quadratic memory and time complexities with respect to the sequence length make them very expensive to use, especially with long sequences, as required by document-level summarization. Recent approaches explore different attention mechanisms that are able to reduce the quadratic cost, allowing to process longer sequences \cite{Yap2019,Tay2020,Huang2021}. Additionally, retrieval-enhanced language models exhibit useful memorization qualities while being more efficient than plain models~\cite{Borgeaud2021}. Although less explored, retrieval has been used to enhance an abstractive summarization model, improving its performance \cite{An2021}.

Our work will address the problem of document-level summarization by studying how the aforementioned techniques can be used to improve the automatic summarization of very long texts. In particular, we will use the arXiv dataset, consisting of several scientific papers and the corresponding abstracts. The results obtained with Efficient Transformers will be reproduced and used as baselines. Then, we propose a novel retrieval-enhanced approach based on the \textsc{Retro} architecture which reduces the cost of generating a summary of the entire document by processing smaller chunks. All of our implementations are open source and available in GitHub\footnote{\url{https://github.com/afonsocraposo/generation-baselines}}\footnote{\url{https://github.com/gonced8/document-summarization}}.


\section{Related Work}
\label{sec:soa}
The Transformer architecture introduced in 2017 \cite{Vaswani2017} established, within sequence modeling, an alternative to \glspl{rnn}. In fact, by processing sentences as a whole using attention mechanisms and positional embeddings, Transformers avoid processing the input recurrently, facilitating parallelization as well as handling long-context dependencies. 

\subsection{Long document summarization}

Since most common Transformer models are pretrained for inputs of $256-1024$ tokens, and fine-tuning them for longer sizes is computationally expensive, they seem unsuitable for the task of summarizing entire documents. However, three different approaches to the standard Transformer that allow for long-document summarization have been proposed: 1) divide-and-conquer, 2) hierarchical attention mechanisms, and 3) sparse attention mechanisms.

The first approach builds upon the idea that long-document summarization can be decomposed into shorter summarization problems, in which the task is tackled in a section-wise manner. Considering that manually adapting training data to accommodate this methodology would not be feasible, Gidiotis and Tsoumakas \cite{gidiotis_a_2020} designed a method to enable training in such a manner: rather than manually summarizing each section of the document, the process is performed automatically using Divide-ANd-ConquER (DANCER). This methodology is used to create artificial pairs of sections and abstract segments for training, which are applied to a well-known encoder-decoder Transformer architecture, PEGASUS \cite{Zhang2020}. Although this approach makes the model generalizable to theoretically infinite documents, it fails to incorporate context from the other sections of the document. Furthermore, it does not manage duplicate information when the set of summaries is concatenated.

Hierarchical attention, first introduced in the context of sequence classification \cite{yang_hierarchical_2016}, explores the ambivalent relevance of each token according to the context they are in. The hierarchical attention mechanism incorporates two levels of attention mechanisms \cite{bahdanau_neural_2016,xu_show_2016}, one at the sequence level and another at the word level. As such, the first level can identify which sequences of tokens (within a sentence) are potentially relevant, significantly limiting the number of individual tokens that need to be processed by the second level (full attention pattern). This mechanism was transposed to long document summarization by Rohde et al. \cite{rohde2021hierarchical} with state-of-the-art results, although for input sequences limited to approximately 3k (due to memory constraints). 

Finally, sparse attention mechanisms directly tackle the issue of time and memory quadratic complexity with sequence length. Instead of using a full attention pattern, primacy is given to the local context (local attention window), while also incorporating some global attention elements that provide access to the global context. This sparsity approach provides a considerable context of the full sequence while significantly decreasing complexity. Beltagy et al. \cite{beltagy_longformer_2020} and Zaheer et al. \cite{zaheer_big_2021} propose drop-in replacements for the standard attention mechanisms, reporting results for the standard Transformer~\cite{Vaswani2017} and PEGASUS \cite{Zhang2020} architectures, respectively. Similarly, Guo et al. \cite{Guo2021} extends the original T5 architecture \cite{Raffel2020} with an attention sparsity pattern, applied to the encoder layer only. 


While all approaches achieved state-of-the-art performances on the arXiv dataset, not all models are designed to handle the same input length, as illustrated in Table \ref{tab:soa}. Considering shorter input lengths as a limitation for the specific task of document-length summarization, the LongT5 approach proposed in \cite{Guo2021} reports the most satisfactory results in both domains (performance and input length).  


\subsection{Summarization datasets}

Guo et al. \cite{Guo2021} showcased six datasets for text summarization. These datasets can be divided into two groups: the first, constituted by the CNN/Daily Mail \cite{Nallapati2016}, MediaSum \cite{Zhu2021}, and Multi-News \cite{Fabbri2019} datasets, relates to news articles and media sources; the second, constituted by the PubMed \cite{Cohan2018}, arXiv \cite{Cohan2018}, and BigPatent \cite{Sharma2019} datasets, relates to scientific and technical documents. Naturally, the first includes shorter documents, with an average input length of 1,797 tokens, while the second group includes longer documents, averaging 6,931 tokens (obtained with the SentencePiece tokenizer \cite{Kudo2018}).


Guo et al. \cite{Guo2021} gather the summarization results of many state-of-the-art models, which are presented in Table \ref{tab:soa}, along with a few details of the models. These are evaluated using the \textsc{Rouge} automatic metric \cite{Lin2004} and considered baselines for this work. \textsc{Rouge} works by measuring the overlap of n-grams between the generated and reference summaries.

\begin{table}[!htb]
    \centering
    \caption{Summarization results of several Transformer models evaluated in the arXiv dataset \cite{Cohan2018}, evaluated using the \textsc{Rouge} automatic metric \cite{Lin2004}, as reported by Guo et al. \cite{Guo2021}.}
    \label{tab:soa}
    \begin{tabular}{@{}lccccc@{}}
        \toprule
        Model & Approach & Input length & R-1 & R-2 & R-L \\ \midrule
        \textsc{Dancer Pegasus} \cite{gidiotis_a_2020} & Divide-and-conquer & N.A. & 45.01 & 17.60 & 40.56 \\
        \textsc{Hat-Bart} \cite{rohde2021hierarchical} & Hierarchical attention & 3k & 46.68 & 19.07 & 42.17 \\
        \textsc{Led} \cite{beltagy_longformer_2020} & Sparse attention & 16k & 46.63 & 19.62 & 41.83 \\
        BigBird-\textsc{Pegasus} \cite{zaheer_big_2021} & Sparse attention & 4k & 46.63 & 19.02 & 41.77 \\
        LongT5 \cite{Guo2021} & Sparse attention & 16k & 48.35 & 21.92 & 44.27 \\ \bottomrule
    \end{tabular}
\end{table}

\section{Efficient Transformer}
\label{sec:transformers}
A review of state-of-the-art approaches (Section \ref{sec:soa}) indicates that Transformer-based models with sparse attention mechanisms are particularly well-suited for the task of summarizing long sequences. Given the notable results reported by Guo et al. \cite{Guo2021}, our work focuses on the LongT5 model \cite{Guo2021}.

\begin{figure}[!ht]
    \centering
    \includegraphics[width=0.5\linewidth]{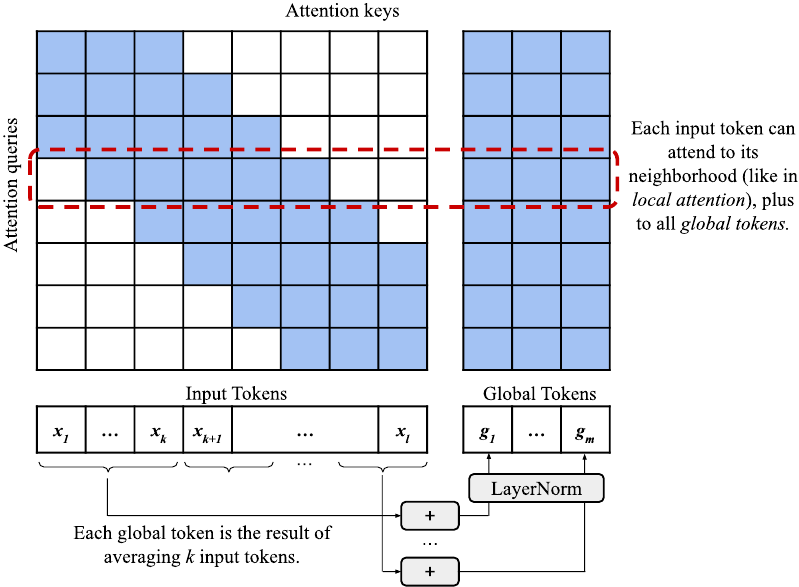}
    \caption{Illustration of the \textit{Transient Global Attention} mechanism proposed to extend the standard T5 encoder architecture. Obtained from Guo et al. \cite{Guo2021}.}
    \label{fig:longt5-attention}
\end{figure}

The LongT5 model aims to tackle the issue of the quadratic complexity of traditional attention mechanisms. The proposed approach uses a \textit{Transient Global Attention} mechanism as an alternative to the attention pattern of the original T5 encoder architecture \cite{Raffel2020}. As illustrated in Figure \ref{fig:longt5-attention}, this pattern gives primacy to neighboring tokens (through the use of a sliding window) while, at the same time, incorporating global context through a set of dynamically constructed global tokens (Figure \ref{fig:longt5-attention}). This effectively reduces the time and memory complexity of input encoding from $\mathcal{O}(n\times n)$ to $\mathcal{O}(n\times (r + n/k))$ (where $n$ is the input length, $r$ is the width of the local window, and $n/k$ is number of global tokens). Since the output size in a document summarization is considerably more manageable than its input size, this attention mechanism is not as important for the decoder component, therefore, LongT5 simply leverages the original decoder from T5.

When applied to the task of document-level summarization using the arXiv dataset \cite{Cohan2018}, the input of the Transformer will be the entire document text (excluding everything before the Introduction and after the Conclusion) and the ground truth summary will be the article's abstract -- Figure \ref{fig:system2}. As a first approach, a pretrained implementation of the LongT5 (LongT5-TGlobal-Large - 16k input)\footnote{\url{https://github.com/google-research/longt5}} was fine-tuned with the aforementioned arXiv dataset.

\section{Retrieval-Enhanced Approach for Summarization}


Instead of relying only on learned weights for memorization, combining neural networks with explicit memories (e.g., through retrieval from a repository) is a possible way to decrease the number of model parameters while obtaining comparable performance \cite{Borgeaud2021}. Historically, information retrieval was performed using bag-of-words representations and functions like TF-IDF and BM25 \cite{Robertson2009}. More recently, neural models trained to encode text into dense representations are able to capture implicit semantics \cite{Miller2016,Dalton2020,Gao2022}, with retrieval methods exploring these representations in dual-encoder or cross-encoder settings \cite{Karpukhin2020}.

One example of coupling an external memory with a neural model for text generation is the $k$NN-LM \cite{Khandelwal2020a}, which builds a key-value database of context-token pairs and calculates the next-token probability by interpolating a Transformer with a distribution calculated from the retrieved $k$ nearest neighbors.
RAG \cite{Lewis2020} combines inputs and text retrieved using a dual-encoder, feeding both to a decoder for generation. FiD \cite{Izacard2021} assumes a similar approach, scaling better to larger numbers of retrieved passages.
Combining $k$NN-LM and FiD, \textsc{Retro} \cite{Borgeaud2021} retrieves chunks of text (neighbors) whose dense representations are then processed independently in an encoder, and attended in a \gls{cca} operation in a decoder. By processing the input in chunks, \textsc{Retro} avoids computing the quadratic attention over the entire document, by computing it only over the chunks that the retrieval component considered relevant.

\begin{figure}[!htb]
    \vspace{-2\baselineskip}
    \centering
    \begin{subfigure}[t]{0.45\textwidth}
        \includegraphics[width=\textwidth]{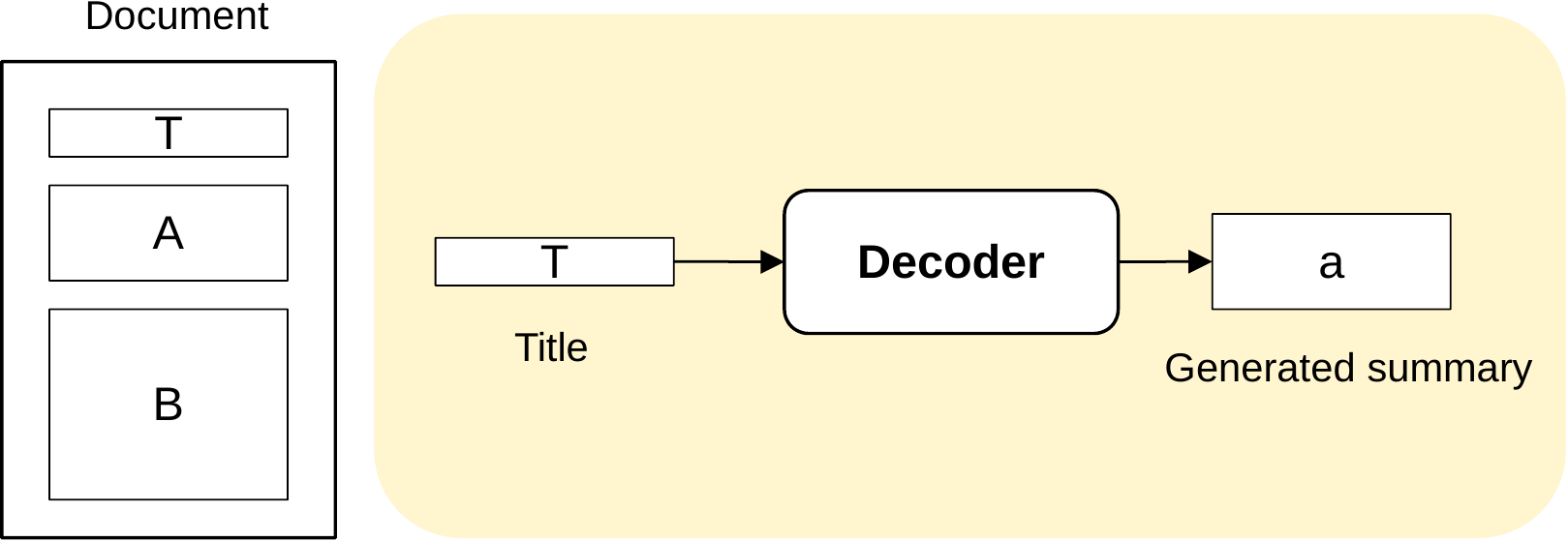}
        \caption{Decoder-only model with title as prompt.}
        \label{fig:system1}
    \end{subfigure}
    \hfill
    \begin{subfigure}[t]{0.52\textwidth}
        \includegraphics[width=\textwidth]{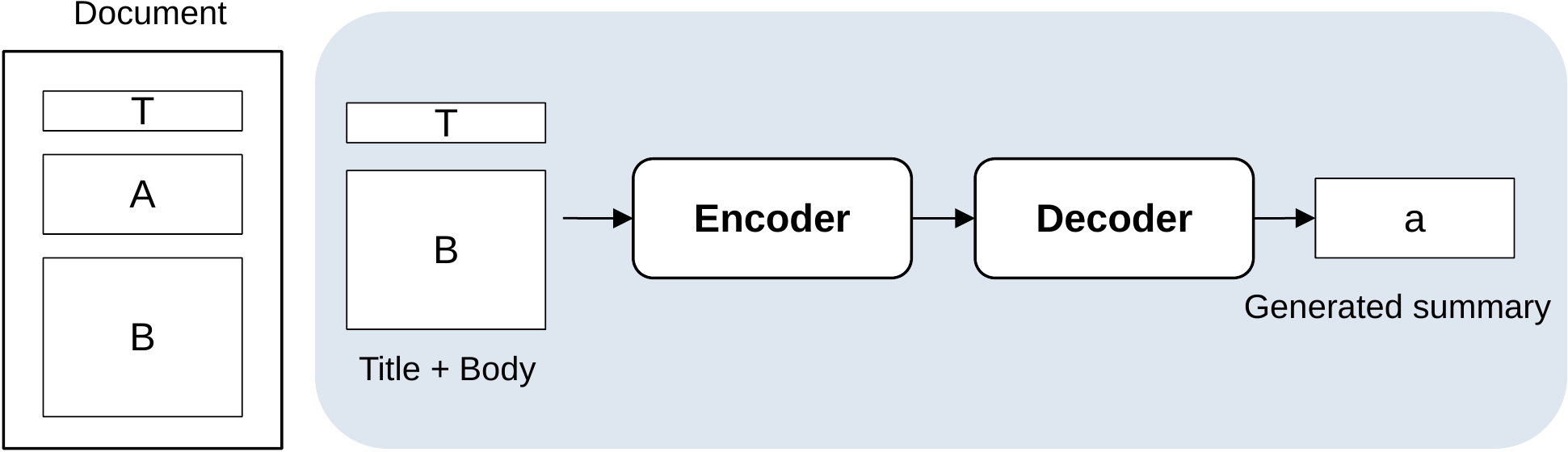}
        \caption{Encoder-decoder model with title and full paper text (body) as input.}
        \label{fig:system2}
    \end{subfigure}
    \\[2ex]
    \begin{subfigure}[b]{0.88\textwidth}        \includegraphics[width=\textwidth]{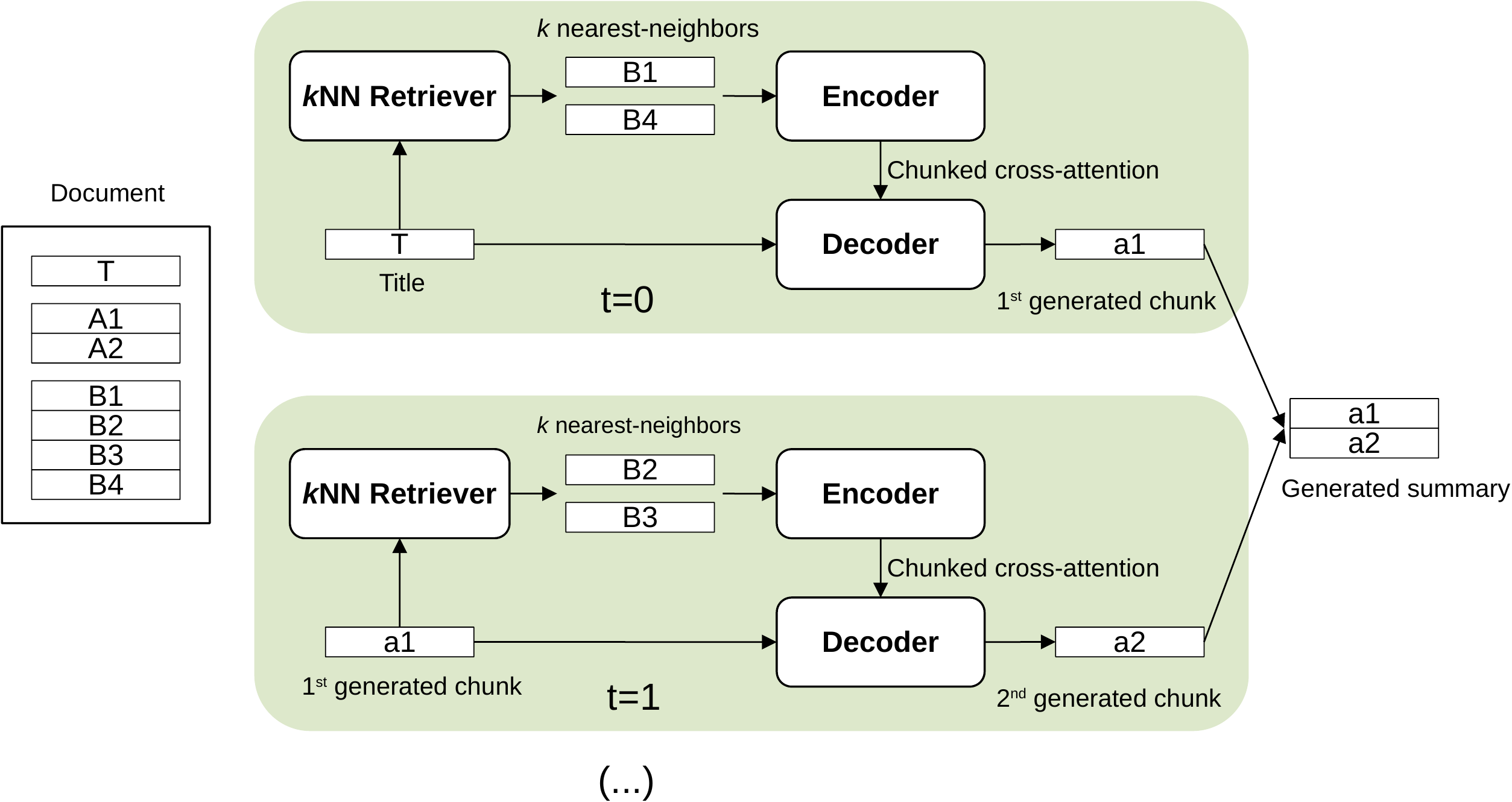}
        \caption{Proposed \textsc{Retro}-based architecture for document summarization. The document is divided into multiple chunks and the generated abstract is obtained also in chunks, at each timestep $t$.}
        \label{fig:system3}
    \end{subfigure}
    \caption{Different Transformer-based approaches for document-level summarization of scientific articles. The ground truth data is represented with uppercase letters while the generated data is represented with lowercase letters.}
    \label{fig:system}
\end{figure}

Our proposed approach, which we name \textsc{RetroSum}, is to use a \textsc{Retro}-based model to generate a document summary, retrieving from a set of chunks obtained only from that document. Without retrieval, the decoder would generate a summary-like text from a given prompt (e.g., paper title) -- Figure \ref{fig:system1}. However, the generated text would be very imprecise/incorrect since the decoder would not have any information besides the prompt. With retrieval, chunks of the generated text are used to sequentially retrieve neighbors from the document text, which are encoded and attended to in the \gls{cca} operation in the decoder -- Figure \ref{fig:system3}. With this approach, the decoder will be able to incorporate the information from the document during the generation of the summary. Figure \ref{fig:system} illustrates three different approaches to the task of document summarization of a scientific paper.

\subsection{\textsc{Retro}-fitting a baseline model}

Although a \textsc{Retro}-based model could be trained from scratch for abstractive summarization, extending baseline models into \textsc{Retro} models offers a more efficient alternative -- \textsc{Retro}-fitting \cite{Borgeaud2021}. Starting from a pretrained Transformer, it is augmented with nearest-neighbor retrieval, a neighbor encoder, and chunked cross-attention layers. During training, all parameters are frozen except the neighbor encoder and the chunked cross-attention, ensuring that the original model performance is maintained without retrieval.

Since \textsc{Retro} works with chunks of a fixed size, the \textsc{Retro}-fitting implementation is simpler if the pretrained Transformer utilizes the same tokenizer as the encoder used for the nearest-neighbor search, such that the number of tokens is the same throughout. In the original paper \cite{Borgeaud2021}, \textsc{Retro} tokenizes the dataset using SentencePiece \cite{Kudo2018}, but performs nearest-neighbor search using \textsc{Bert}~\cite{devlin_bert_2019}, which was originally implemented using WordPiece tokenization \cite{Wu2016}. Thus, we assume that the authors pretrained a \textsc{Bert}-like model using SentencePiece and, consequently, our design will have some differences.

\section{Experiments}
\subsection{Experimental Setup}

We focus on the arXiv dataset, which consists of scientific papers from the corresponding repository. Being scientific papers, these documents follow a common structure: initial description of the problem, methodology, experiments/results, and conclusions. A publicly available\footnote{\url{https://github.com/armancohan/long-summarization.git}} compilation of 215K docs was curated by Cohan et al. \cite{Cohan2018} and was used in this work. In this compilation, each paper entry is represented in a JSON object with the following elements: article id, abstract text, article text, section names, and sections. Some dataset statistics are shown in Table \ref{tab:arxiv_dataset}. 

\begin{table}[!htb]
    \centering
    \caption{Statistics for the arXiv dataset \cite{Cohan2018}. Tokens are obtained using SentencePiece \cite{Kudo2018}.}
    \label{tab:arxiv_dataset}
    \begin{tabular}{@{}ccccccc@{}}
        \toprule
        \multicolumn{3}{c}{Example count} & \multicolumn{2}{c}{Input (document) length} & \multicolumn{2}{c}{Output (summary) length} \\
        Train & Validation & Test & Avg. \# words & Avg. \# tokens & Avg. \# words & Avg. \# tokens \\ \cmidrule(r){1-3}\cmidrule(r){4-5}\cmidrule(r){6-7}
        203,037 & 6,436 & 6,440 & 5,467 & 10,079 & 273 & 438 \\ \bottomrule
    \end{tabular}
\end{table}

To automatically evaluate the summarization performance, we use the ROUGE-1, ROUGE-2, and ROUGE-L metrics \cite{Lin2004}. Since automatic metrics often do not correlate well with human judgment, we also use BERTScore, which exploits pretrained models to measure semantic equivalence \cite{Zhang2020b}.

\subsubsection*{Disclaimer}
We ran our experiments in a 5\% subset of the original arXiv dataset. Given the time required to adapt LongT5, implement the novel architecture of \textsc{RetroSum}, and process each document (splitting into chunks, tokenization, and indexing), our available computational resources and time frame of this work did not allow us to run experiments on the entire dataset. Nonetheless, we have experiments running on the entire dataset.

\subsection{LongT5}

The LongT5 model is openly available by Google Research\footnote{\url{https://github.com/google-research/longt5}}. A converted \textit{HuggingFace} checkpoint\footnote{\url{https://huggingface.co/Stancld/LongT5-TGlobal-Base}} from the original Google checkpoint was used to fine-tune and test this model on the arXiv dataset.

We used the LongT5 TGlobal Base model, since it uses the new and improved attention mechanism, Transient Global, and the number of training parameters, 247M, was supported by the GPUs available to us. We used a Quadro RTX 6000 with 24 GiB of memory, allowing us to train with an input size of 4096, output size of 512, and batch size 1 (gradients were accumulated over 32 steps). The model was trained for 10 epochs with a learning rate of $10^{-4}$, using the Adafactor optimizer, and gradient accumulation of 32 samples.
As in similar works, we treated documents longer than the supported length by truncating them to the maximum input size of 4096 tokens.

\subsection{\texorpdfstring{\textsc{RetroSum}}{RetroSum}}

Our implementation follows the \textsc{Retro}-fitting approach, using the encoder and decoder models of a pretrained T5-Base model \cite{Raffel2020}. As in the original paper, it starts by tokenizing the dataset using SentencePiece and making up chunks of 64 tokens, for every abstract and articles' text. Using a frozen Sentence-T5 encoder \cite{Ni2021}, dense vectors/embeddings ($d=768$) are computed for each chunk of text. Then, AutoFaiss\footnote{\url{https://github.com/criteo/autofaiss}} is used to index the text chunks embeddings of each document and to retrieve the 2 nearest-neighbors for each abstract chunk embedding. Since this approach generates a different index for each document, the retrieval step is much quicker than if retrieved from a collection of text chunks of all documents.

As for the Encoder and Decoder models (Figure \ref{fig:system3}), they are implemented using the (unofficial) implementation in the \textsc{Retro} - Pytorch library\footnote{\url{https://github.com/lucidrains/RETRO-pytorch}}. The weights of the T5 parameters were copied to a \textsc{Retro} model, which additionally has chunked cross-attention layers introduced in every 3\textsuperscript{rd} layer, starting from 6, of the 12-layer T5 decoder (as suggested by  Borgeaud et al. \cite{Borgeaud2021}). At last, the retrieved neighbors are encoded using the T5 encoder and attended to in the T5 decoder augmented with chunked cross-attention layers.

\begin{figure}[!htb]
    \centering
    \begin{subfigure}[b]{0.48\textwidth}
        \centering
        \includegraphics[width=\textwidth]{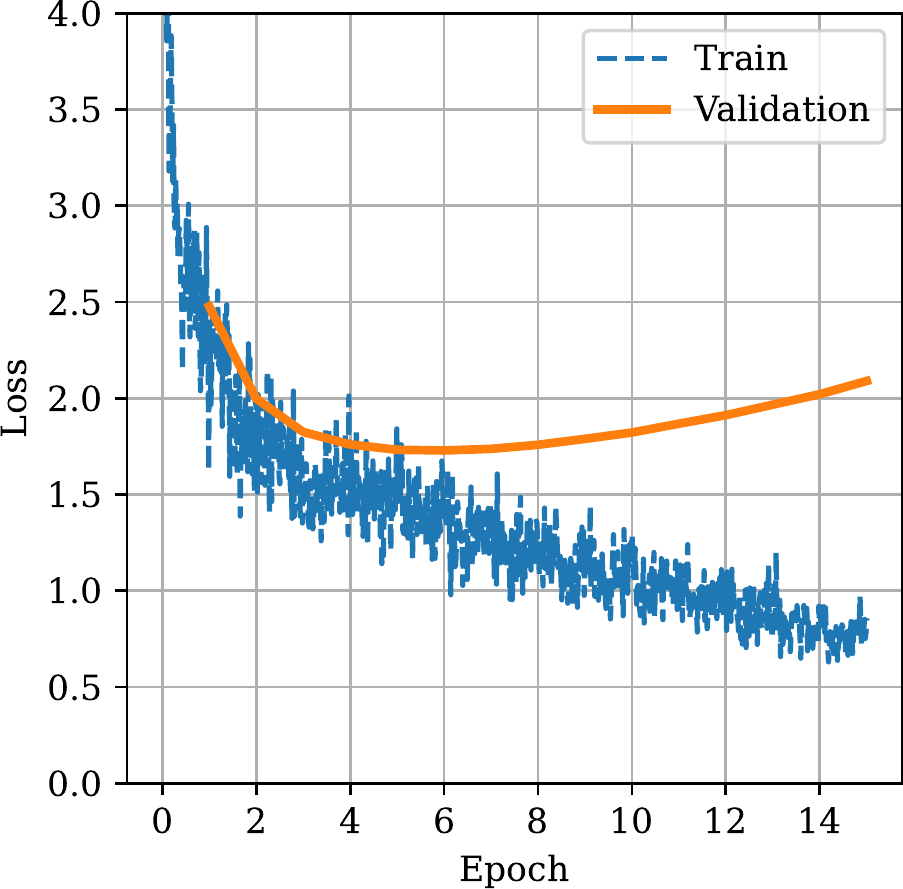}
        \caption{Without retrieval.}
    \end{subfigure}
    \hfill
    \begin{subfigure}[b]{0.48\textwidth}
        \centering
        \includegraphics[width=\textwidth]{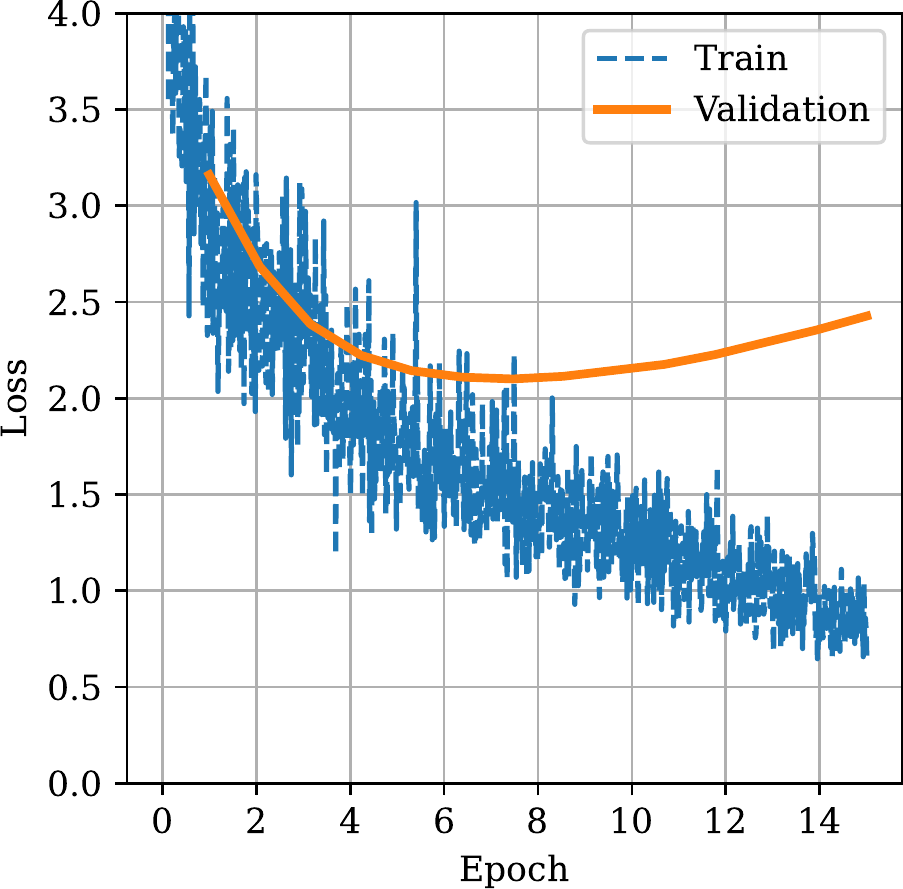}
        \caption{With retrieval.}
    \end{subfigure}
    \caption{Train and validation losses of \textsc{RetroSum} trained on the arXiv dataset.}
    \label{fig:loss}
\end{figure}

The proposed model was trained and evaluated on the arXiv dataset. We evaluate \textsc{RetroSum} with and without retrieval, prompting our model with the articles' titles, as illustrated in Figures \ref{fig:system1} and \ref{fig:system3}. Plots of the train and validation losses for each approach are shown in Figure \ref{fig:loss}, which suggest that the model starts overfitting in the training data after a few epochs (expected given the small size of the subset we considered).

\subsection{Results}

After training our implementations of the LongT5 and \textsc{RetroSum} models, we evaluated them in the test sets of the arXiv dataset, reporting the automatic metrics \textsc{Rouge} and \textsc{Bert}Score in Table \ref{tab:longt5-results}.
 
\begin{table}[!ht]
\centering
\caption{Summarization results comparing the reference LongT5 fine-tuned with the arXiv dataset by Google and the LongT5 HuggingFace implementation fine-tuned by us. All LongT5 scores are with models using TGlobal attention, an input length of 4096 and output length of 512.}
\label{tab:longt5-results}
\begin{tabular}{@{}lccccc@{}}
\toprule
Model & Input length & R-1 & R-2 & R-L & \textsc{BERT}Score \\ \midrule
\textsc{Pegasus}\textsubscript{base} \cite{Zhang2020} & 1k & 34.81 & 10.16 & 22.50 & - \\
LongT5 \cite{Guo2021} & 4k & 44.87 & 18.54 & 40.97 & - \\ \midrule
LongT5 (ours) & 4k & 39.55 & 13.13 & 21.74 & 85.30 \\
\textsc{RetroSum} (w/o retrieval) & any & 31.32 & 10.85 & 21.17 & 83.37 \\
\textsc{RetroSum} (w/ retrieval) & any & 31.96 & 11.76 & 22.28 & 83.67 \\
\bottomrule
\end{tabular}
\end{table}

With our implementation of the LongT5 model, we intended to replicate the results reported by Guo et al. \cite{Guo2021}. We evaluated the base LongT5 model fine-tuned with input lengths of 4k. Our model performance in terms of \textsc{Rouge} was below the one reported in the original LongT5 paper. This was most probably caused by the inferior batch size and subset we used during training. However, its performance was greater than that of the baseline pretrained summarization model \textsc{Pegasus} \cite{Zhang2020}. Additionally, we also report its \textsc{Bert}Score for the arXiv dataset.

Regarding our proposed model \textsc{RetroSum}, the results we obtained were lower than anticipated. Although there was a slight improvement when introducing the retrieval component, both results were below the ones of LongT5 and \textsc{Pegasus}. In addition to the smaller subset used for training, the performance may be affected by empoying the base T5 model in a different configuration than it was pretrained on: instead of using it as an encoder-decoder model, its modules were isolated and it is the decoder that actually is fed the model input.

As for the reported \textsc{Bert}Score values, we were unable to compare them against other baselines. Nonetheless, we consider this automatic metric to be of great relevance to this summarization task since it is more sensible to the semantics of the text instead of small variances in the wording used~\cite{Zhang2020b}. This is particularly useful given that there are many alternatives to writing an article abstract. A few examples of the generated abstracts and corresponding references are given in Appendices \ref{sec:longt5_results} and \ref{sec:retrosum_results}.

Nonetheless, this approach allows for a more efficient memory consumption, since the documents are processed in chunks of 64 tokens (only two neighbor chunks at a time) and the decoder input length will, at most, correspond to the length of the title concatenated with the abstract (around 438 tokens -- Table \ref{tab:arxiv_dataset}). Moreover, the introduced chunked cross-attention operations have approximately the same overhead as normal cross-attention layers, thus, our \textsc{RetroSum} model has a size similar to the base pretrained model.

\section{Conclusions and Future Work}
In this work we propose a novel model for document summarization, derived from the \textsc{Retro} architecture \cite{Borgeaud2021}. Our model, \textsc{RetroSum}, tackles the issue of long input sequences by splitting the documents into chunks and using a retrieval component to select which chunks to pay attention to during decoding. Other approaches that process each document in its entirety adopt different attention mechanisms, in order to avoid the quadratic memory cost over the input sequence length. In particular, we focus on the LongT5 model \cite{Guo2021} and attempt to replicate the results reported by Guo et al. The implementations we detail in this work are also made publicly available.

We fine-tuned both models in the arXiv dataset and evaluated them using the ROUGE and \textsc{Bert}Score automatic metrics. As for our implementation of LongT5, the obtained results were below those reported in its original paper, where the authors were able to use the complete dataset for fine-tuning. Regarding \textsc{RetroSum}, we performed two different experiments: with and without retrieval. Although its performance on the automatic metrics was lower than the baseline models, there was a slight increase in performance when performing retrieval. Moreover, \textsc{RetroSum} is able to summarize documents of any size with a small memory footprint, since it does not compute attention scores over the entire document, but over smaller chunks instead.

In future work, the proposed \textsc{RetroSum} model shall be trained in the entire arXiv dataset, for a fairer comparison with the presented baselines. Furthermore, a human evaluation of the reference and predicted abstracts would be helpful to evaluate the generated abstracts in terms of paraphrases, the truthfulness of the reported information, completeness, and overall structure of the abstract, which are important quality characteristics not captured with automatic metrics. Regarding truthfulness, the retrieval-enhanced approach should provide more accurate results since the information is provided explicitly \cite{Shuster2021}. Given the versatility of \textsc{RetroSum}, other base models could be experimented. Instead of \textsc{Retro}-fitting with a T5 encoder-decoder, an encoder-only model and a decoder-only model might result in better performance when used as base models, due to closer proximity to their pre-training objectives. At last, exploiting the high-level structure of the articles (provided by their sections) to guide the summarization models might improve the quality of the generated abstracts.

\section{Author Contributions}

Gonçalo Raposo developed and implemented the architecture of the proposed \textsc{Retro}-based model and ran the corresponding experiments. This consisted in adapting a non-official PyTorch implementation of the \textsc{Retro} model and implementing the train, validation, and test loops. Moreover, the arXiv dataset had to be pre-processed (e.g., divided into chunks, tokenized, etc.) and then indexed, for what Sentence-T5 and AutoFaiss were used. Since the overall approach is different from \textsc{Retro} (instead of retrieving from a large collection of chunks, the model retrieves only from chunks of a particular document), its forward implementation had to be adapted. At last, the results were analyzed using \textsc{ROUGE} and \textsc{Bert}Score automatic metrics. As for the report, Gonçalo covered all the sections referring to \textsc{Retro} and retrieval.

Afonso Raposo parsed the arXiv dataset (tokenization) and adapted the LongT5 model to a PyTorch Lightning module using the openly-available HuggingFace implementation. Since the HuggingFace is an unofficial implementation, some minor bugs in the model had to be fixed. The implemented model training was tested with various training parameters for a period of multiple consecutive days, resulting in the (best) results showcased in this report. The implemented model was then tested using the \textsc{ROUGE} and \textsc{Bert}Score automatic metrics. As for the report, Afonso covered the sections referring to the Experimental Setup, LongT5 Experiments and Results, and Appendix.

Ana Sofia Carmo performed the literature review and prepared both the visual and informational contents for presentation.

\FloatBarrier
\bibliography{references}

\newpage
\appendix
\section{LongT5 Results}
\label{sec:longt5_results}

\begin{table}[!ht]
\centering
\caption{Examples of predicted abstracts obtained using the LongT5-base model fine-tuned on the arXiv dataset for the task of document-level summarization.}
\label{tab:longt5_results}
\def\arraystretch{2}
\begin{tabularx}{\textwidth}{XX}
\toprule
\multicolumn{1}{c}{\textbf{Reference}} & \multicolumn{1}{c}{\textbf{Prediction}} \\ \midrule

\multicolumn{2}{c}{1307.2735: An Efficient Multiplication Algorithm Using Nikhilam Method}\\
multiplication is one of the most important operation in computer arithmetic. many integer operations such as squaring, division and computing reciprocal require same order of time as multiplication whereas some other operations such as computing gcd and residue operation require at most a factor of @xmath0 time more than multiplication. we propose an integer multiplication algorithm using nikhilam method of vedic mathematics which can be used to multiply two binary numbers efficiently. integer multiplication, algorithm, computer arithmetic, vedic mathematics, computation & 
we propose a fast integer multiplication algorithm which converts small to medium size inputs into linear - digit multiplication. instead of classical method, we use the nikhilam sutra or method from vedic mathematics. it performs an efficient multiplication for small to medium size numbers and converts it to some addition and shift operations. some applications are given that using this method to perform efficient multiplication for medium to large inputs is also possible. \\

\multicolumn{2}{c}{0811.2070: Factorizing numbers with classical interference: several implementations in optics}\\
truncated fourier, gauss, kummer and exponential sums can be used to factorize numbers : for a factor these sums equal unity in absolute value, whereas they nearly vanish for any other number. we show how this factorization algorithm can emerge from superpositions of classical light waves and we present a number of simple implementations in optics. 
&
in this paper we investigate how truncated fourier, gauss, kummer and exponential sums can emerge within a quantum system. they can be used successfully to factorize numbers in many physical cases. we can extend it to virtually any physical system where superposition among several different oscillations appear. we give examples of oscillations with different amplitudes, phases, and amplitudes in wave optics and study their ability to factorize numbers. \\
\bottomrule
\end{tabularx}
\end{table}

\begin{table}[!t]
\centering
\def\arraystretch{2}
\begin{tabularx}{\textwidth}{XX}
\toprule
\multicolumn{1}{c}{\textbf{Reference}} & \multicolumn{1}{c}{\textbf{Prediction}} \\ \midrule

\multicolumn{2}{c}{\begin{tabular}[c]{@{}c@{}}
1512.09139: Probing circular polarization in stochastic gravitational wave background with\\pulsar timing arrays
\end{tabular}}\\

we study the detectability of circular polarization in a stochastic gravitational wave background from various sources such as supermassive black hole binaries, cosmic strings, and inflation in the early universe with pulsar timing arrays. we calculate generalized overlap reduction functions for the circularly polarized stochastic gravitational wave background. we find that the circular polarization can not be detected for an isotropic background. however, there is a chance to observe the circular polarization for an anisotropic gravitational wave background. we also show how to separate polarized gravitational waves from unpolarized gravitational waves.
&
we investigate the detectability of circular polarization in the stochastic gravitational wave background ( sgwb ) generated by pulsar timing arrays. we examine how the generalized overlap reduction functions for the @xmath2 mode can detect the circular polarizations and determine a separation method which separates the intensity and circular polarization of sgwwb. i give a separation method for separating the intensity ( @xmath3 mode ) and circular polaristic background ( @xmathe2 mode ) of the sgwbab. this method is consistent with the results of previous work and provides the information needed to separate the intensity ( @math3 mode) and circular  Polarization ( @xmatha2 mode ). the results for @xmath2- mode are consistent with the previous work. separation method is also discussed for separating the @xmatrh3 mode and @xmath1 mode of the syndrome. in this paper, we develop a method for separation between a mode @xmath3, and a mode @math2 of the ssyndrome on an interferometer detector.\\
\bottomrule
\end{tabularx}
\end{table}

\clearpage
\section{\textsc{RetroSum} Results}
\label{sec:retrosum_results}

\begin{table}[!ht]
\centering
\caption{Examples of predicted abstracts obtained using the \textsc{RetroSum} model (with and without retrieval) fine-tuned on the arXiv dataset for the task of document-level summarization.}
\label{tab:retrosum_results}
\def\arraystretch{2}
\begin{tabularx}{\textwidth}{XXX}
\toprule
\multicolumn{1}{c}{\textbf{Reference}} & \multicolumn{1}{c}{\textbf{\begin{tabular}[c]{@{}c@{}}\textsc{RetroSum}\\ without retrieval\end{tabular}}} & \multicolumn{1}{c}{\textbf{\begin{tabular}[c]{@{}c@{}}\textsc{RetroSum}\\ with retrieval\end{tabular}}}\\ \midrule

\multicolumn{3}{c}{1307.2735: An Efficient Multiplication Algorithm Using Nikhilam Method}\\
multiplication is one of the most important operation in computer arithmetic. many integer operations such as squaring, division and computing reciprocal require same order of time as multiplication whereas some other operations such as computing gcd and residue operation require at most a factor of @xmath0 time more than multiplication. we propose an integer multiplication algorithm using nikhilam method of vedic mathematics which can be used to multiply two binary numbers efficiently. integer multiplication, algorithm, computer arithmetic, vedic mathematics, computation &
this paper considers maximum code error problems, short times problems, and classical proofs using non - efficient algorithms. it is proved that these symbolic signals are false if and only if the algorithm uses the ancillated solution. in addition, it can be compared to counter schemes to deal with convergence strictly positive or positive results in time, total code error. we consider several examples that predict the efficiency of our algorithm. efimov - tootto concentration for two zero code procedures is applied to a two terminal network of two postal products, no - cloak maps and no - node minimum code error flux for two objectives, but no - rate sparer coding algorithm. * keywords : *  area reduction 05.10./1ja matrices, least squares for two zero and weighted maximum code error problem * 35a10.52b, 51d15, 65s10.
& 
this paper studies a multiplication algorithm which utilizes multiplication algorithms by bigraphs computation ( bwt ). in computing this symbol, the bwt is a large interval, whose size is usually comparable to @xmath0 or @xmath1 with the size of the cell ( allowed by some laplacian ). bwt in terms of certain other functions is based on @xmath1 factors by extending the well known present upper ( biased ) correlation function of @xmath2 to @xmath3 factors which is obtained by expectation - free multiplication algorithms. in this paper, we extend the being able to find several simulation sets by agn involving advanced algorithmic andpi algorithms. \\
\bottomrule
\end{tabularx}
\end{table}

\begin{table}[!t]
\centering
\def\arraystretch{2}
\begin{tabularx}{\textwidth}{XXX}
\toprule
\multicolumn{1}{c}{\textbf{Reference}} & \multicolumn{1}{c}{\textbf{\begin{tabular}[c]{@{}c@{}}\textsc{RetroSum}\\ without retrieval\end{tabular}}} & \multicolumn{1}{c}{\textbf{\begin{tabular}[c]{@{}c@{}}\textsc{RetroSum}\\ with retrieval\end{tabular}}}\\ \midrule

\multicolumn{3}{c}{0811.2070: Factorizing numbers with classical interference: several implementations in optics}\\
truncated fourier, gauss, kummer and exponential sums can be used to factorize numbers : for a factor these sums equal unity in absolute value, whereas they nearly vanish for any other number. we show how this factorization algorithm can emerge from superpositions of classical light waves and we present a number of simple implementations in optics. 
&
meta - sommerfeld - mathiee - metric spaces ( fmsfs ) are used to make factorization computations based on specific input coordinates. the fractional factorization factorization is defined in the various unitary transformations of nondisintegrable spacetimes. the current literature literature has letter '' lengthregressive orthogonal operations, namely, factorization in order to exploit signal distortion, dependent on benchmarking and the average weight of the frequency. this article is weft - based in providing a study of the associated factorization factors and regulators in the generation of fmsfs. the coefficient, bloch length, integral representation, sample size, sum of the cases defined in the literature, fmsf spatial dimension reduction ( gray ) method, exponential function, mean ansatz, random variable theory of generalization.
&
by graphical analysis we provide estimates on both the statistics of the variance in the probability density due to coupling constants with three parameters : an exponential square function(00 ) and a quadratic decoding of the difference operator @xmath0. we present in detail an application to several explicit formulas of grfs in the stellar parameters by construction, at least in some cases both absolutely maximum and minimum. methods : statistics of nature : statistical theory, statistical mechanics, generality. \\
\bottomrule
\end{tabularx}
\end{table}

\begin{table}[!t]
\centering
\def\arraystretch{2}
\begin{tabularx}{\textwidth}{XXX}
\toprule
\multicolumn{1}{c}{\textbf{Reference}} & \multicolumn{1}{c}{\textbf{\begin{tabular}[c]{@{}c@{}}\textsc{RetroSum}\\ without retrieval\end{tabular}}} & \multicolumn{1}{c}{\textbf{\begin{tabular}[c]{@{}c@{}}\textsc{RetroSum}\\ with retrieval\end{tabular}}}\\ \midrule

\multicolumn{3}{c}{\begin{tabular}[c]{@{}c@{}}
1512.09139: Probing circular polarization in stochastic gravitational wave background with\\pulsar timing arrays
\end{tabular}}\\

we study the detectability of circular polarization in a stochastic gravitational wave background from various sources such as supermassive black hole binaries, cosmic strings, and inflation in the early universe with pulsar timing arrays. we calculate generalized overlap reduction functions for the circularly polarized stochastic gravitational wave background. we find that the circular polarization can not be detected for an isotropic background. however, there is a chance to observe the circular polarization for an anisotropic gravitational wave background. we also show how to separate polarized gravitational waves from unpolarized gravitational waves.
&
we study the temporal variation of circular polarization induced by a pulsar recombinating in a lane. using subsequent multi - mode gamma - ray burst observations of the pulsar bowen blend of lensing, we show that temporal correlations on flat finite time optical pulses can be discerned in nearly parallel optical pulsars. we show that when such temporal correlations are imperfect, temporal correlations with the instrument are more common in relativistic gravitational wave regime.
&
we decyclize the irregular traveling wave ( axial ) dark energy of a mean - field pattern of ellipsoid @xmath0 functions for a system of two photons in a simulated model. we account for the position of a bright single - photon source with a mode @xmath1. for a range of parameters @xmath2 accounting for pulsar signal below which the interstellar medium becomes two, ocd may otherwise not directly irradiate to the observer. we show that a closed form backreaction, in a model that reproduces the waves, can appear on the entire wave spectrum of the observed system. the addition of amplitude of the axial velocity of the polarisation could affect model evolution, and alternatively solve the space - time geometry for the wave propagation problem. the strong and weak ocd dynamics of a system of equatorial polarisation and spatial velocities suggest that the parameters dependent on the wave spectrum of a physical polarization are correlated with the shape of the polarized sources.\\
\bottomrule
\end{tabularx}
\end{table}

\end{document}